\documentclass[journal]{Template/IEEEtran}
\usepackage{graphicx}
\usepackage{siunitx}
\usepackage{cite}
\usepackage{gensymb}
\usepackage{comment}
\usepackage[figurename=Fig.,font=scriptsize]{caption}
\usepackage{subfig}

\begin{document}
\title{Studying Accuracy of Machine Learning Models Trained on Lab Lifting Data in Solving Real-World Problems Using Wearable Sensors for Workplace Safety}
\author{Joseph Bertrand, Nick Griffey, Ming-Lun Lu, Rashmi Jha}%
\maketitle

\begin{abstract}
Porting ML models trained on lab data to real-world situations has long been a challenge. This paper discusses porting a lab-trained lifting identification model to the real-world. With performance much lower than on training data, we explored causes of the failure and proposed four potential solutions to increase model performance.
\end{abstract}

\section{Introduction}
\label{intro}
\IEEEPARstart{B}ack pain is a leading cause of occupation-induced disability and results in large amounts of lost productivity annually \cite{back_pain_leading_cause_of_injury}. Most occupational back pain is caused by roles requiring repetitive tasks such as heavy lifting of objects. There are many ergonomic guidelines for lifting with the goal of reducing the risk of back pain, but it is difficult for workers to consistently follow these guidelines in all situations. Many may be unable to lift in the required manner due to poor workstation designs or physical limitations \cite{poor_workstation_design}. Monitoring a lifting workstation can ensure compliance with safe lifting techniques as well as aid in determining if workers are consistently lifting in an unsafe manner to perform the work the job requires. However, this type of monitoring requires significant overhead and doesn't provide sufficient data to create an objective methodology to assess risk. An automated system would allow users to classify risk with minimal overhead and provide real-time feedback to workers in an attempt to reduce long-term risk.\\

Both lift classification and lift detection are known as human activity recognition (HAR), which have been extensively studied in machine learning under various sensor modalities \cite{extensively_studied}. HAR systems have been successful in video-based deployments \cite{video_success}. However, automatic lift assessment would require cameras to be placed in every lift location, which is prohibitively expensive and raises concerns for the privacy of the workers. Furthermore, it is impractical for temporary workstations such as construction zones and contract work. In contrast, Inertial Measurement Unit (IMU) sensors are wearable, can detect motion regardless of workplace location, and don't have the privacy concerns associated with cameras. IMU sensors are widespread in common consumer devices, such as smart watches and phones and therefore provide significant advantage for HAR problems.\\

Previous work was done to develop a model that could identify a lifting event from a subset of laboratory-gathered data with an F1 score of ~97\% \cite{brennans_paper}. F1 is the harmonic mean of precision and recall and is commonly used to evaluate binary classifiers. The evaluation data was randomly sampled using K-fold Cross Validation. The model trained was able to reliably identify lifting and non-lifting events from data gathered under the same conditions as its training data. However, when this model was applied to a real-world environment (data collection in lab and real world is discussed in section II), the F1 score dropped to 32.8\%, showing that while the model could reliably identify lifting events within a dataset environment, it failed to identify more general lifting events. \\

Poor general performance on a model is not uncommon, and there is a significant amount of general guidance published\cite{overfitting_paper}. However, most of this work pertains to datasets where the capture device use does not significantly affect the data. For example, creating an object recognition model from images, the camera settings do not majorly change the underlying dataset, since all reasonable cameras produce clear images, either with or without the relevant recognition object. However, in HAR models, the location of the IMU sensor majorly changes the data, and it is not currently possible to standardize IMU data from sensors placed in different locations on the body. This makes it unfeasible to use some common techniques for improving model performance, such as transfer learning and generalizing training data from public datasets. Therefore, adjustments to previous techniques were developed to make them suitable for HAR model applications.

\section{Background}
\label{background}

\subsection{Data Collection} 
\begin{figure}[h]
    \centering
    \includegraphics[width=0.40\textwidth]{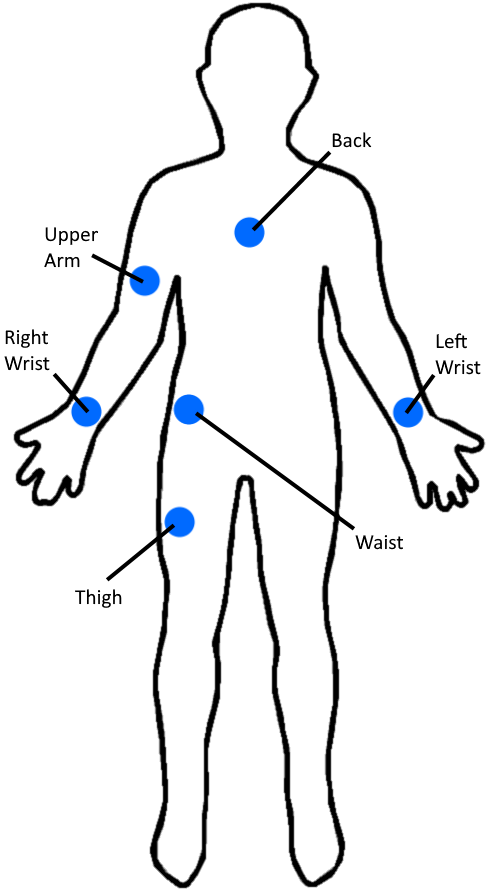}
    \caption{Placement of IMU sensors on subjects}
    \label{fig:sensor-locations}
\end{figure}

To gather data, each subject had six IMU sensors attached to various locations on their body. One sensor on each wrist, one on the upper-right thigh, one on the upper back, one on the upper-right arm, and one on the waist (figure \ref{fig:sensor-locations}). Each IMU sensor gathered accelerometer and gyroscopic data at a sampling rate of 25hz.\\ 

\subsubsection{Lab Data}
Lab data was gathered in Phase 1 \& 2 of the NIOSH research (IRB approved study protocol: 16-DART-05XP). Therefore, it may be referred to as phase 1\&2 data throughout this paper. To gather the data, each subject started in the center of a room. They would then walk forward and pick up a small, wired box (this denoted the start of the lift). They would then turn and walk with the box across the room and set the box down on a table. Setting the box down denoted the end of the lifting event. After they set the box down, they would turn and walk back to the center of the room, where the trial would stop.\\

NIOSH recorded videos of each trial and used a motion capture system (MoCap) to accurately track the location of the box. The start and end lift times were determined from the motion capture system. To identify the start and stop lift time for training and testing the IMU sensor model, code was written to synchronize the IMU data with the MoCap data. The time of the lift, in hh:MM:ss:ms format, was stored, from MoCap data, as metadata for each lift trial and each IMU sensor data point had a Unix Epoch timestamp associated with it. To synchronize the two, code was written to extract the time of day from the Unix timestamp, then using the time of day from the metadata, and the knowledge that the sampling rate of the IMUs was 25HZ, we could calculate which IMU frame the lift started and ended. The calculations and code were manually verified.\\

\subsubsection{Real-World Data}
Real-world data was gathered in Phase 3 of the NIOSH research (IRB approved study protocol: 16-DART-05XP). Therefore, it may be referred to as phase 3 data throughout this paper. Phase 3 data was gathered in the field at a light-industrial plant. Subjects wore the sensors in the same arrangements as they did in the phase 1\&2 data. Two Sony Handycam digital cameras were placed at two angles of view (side and front view) of the subject to allow the NIOSH team to review and manually label the beginning and ending of each lift. The labeled lifts were used to assess the model in the evaluation stage (See section V). There were some points where certain sensors were non-functional. That data was removed from the dataset to maintain consistency with the expected six-sensor data format.\\

To perform model evaluation, the team sampled a small segment of phase 3 data and used subjective analysis to determine if the model was performing well. As model development continued, the team decided to go back and label the phase 3 dataset so models could be compared objectively. After balancing the dataset there were approximately 950 seconds of data used for evaluation.

\subsection{Identification Model Training}

For the purposes of this paper the beginning of the lifting motion (BOL) was defined by when the object that the subject was lifting began to move, and the end of the lifting motion (EOL) was defined by when the subject had fully lifted the object and had either placed it down, or brought it close to their body. We defined lifting with these parameters because the majority of the risk associated with the lifting motion is within these two time intervals \cite{niosh_lifting_eq} The lifting labeling in Phase 1\&2 training data differed slightly from this definition. For the NIOSH labeling, two researchers reviewed the trials and identified the frame number where the grid started to move (Beginning of Lift) and where the grid was set down completely by two hands (End of Lift). If there was any doubt, the researchers would discuss and reach an agreement on the final frame \cite{niosh_labeling_paper}. In order to fit the NIOSH labeling to the definition of lifting used in this project, the EOL was adjusted to be 1.52 seconds after the BOL. This time was chosen by reviewing the lifting videos and finding the average time it took for a subject to complete the core lifting motion. For the evaluation dataset, the beginning of lift and end of lift labels provided by NIOSH matched our lifting definition very closely, so no adjustments were necessary. \\

To train the lift identification model, the phase 1\&2 dataset was divided equally into two categories, lifting events and non-lifting events. Since the raw dataset contained more non-lifting events than lifting events, the dataset was balanced randomly to include roughly the same number of both classes. The lifting event was a 1.2 second window, starting from the beginning of lift index provided by NIOSH. 1.2 seconds was the approximate average lifting time based on examining videos of the subject's lifts. Everything outside that 1.2 second window was considered to be a non-lifting event. There were also several files provided which contained no lifting events, and instead contained events such as walking, sitting, or jumping. Those files were also considered to be non-lifting events.\\

Once a balanced dataset was generated, the model was trained using keras's .fit method. The hyperparamaters batch size, feature length, epoch, and validation split, were varied in model in an attempt to find the combination that provided the best performance. A total of 3456 models were trained, using a combination of various predefined hyperparamater combinations. All models had the same neuron architecture, consisting of an input layer, followed by a 128-layer deep LSTM layer, followed by two Dense functions of width 5, then with an dense output layer of size 1 to provide the final result. This model architecture was evaluated on phase 1\&2 data and had performance parity with previous work \cite{brennans_paper}. \\

Each model was then evaluated on phase 3 data. The evaluation consisted of obtaining the accuracy, F1, and confusion matrix from each model. The results were cataloged to create the results data seen below in the results section.

\section{Challenge Points}

\subsection{Data Investigation}
\subsubsection{Data Offset}
\begin{figure}[h]
    \centering
    \includegraphics[width=0.49\textwidth]{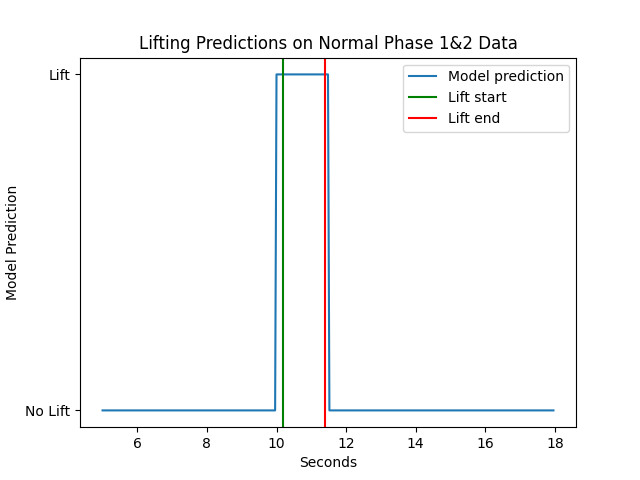}
    \caption{Typical lifting predictions on Phase 1\&2 data}
    \label{fig:accurate_predictions}
\end{figure}

\begin{figure}[h]
    \centering
    \includegraphics[width=0.49\textwidth]{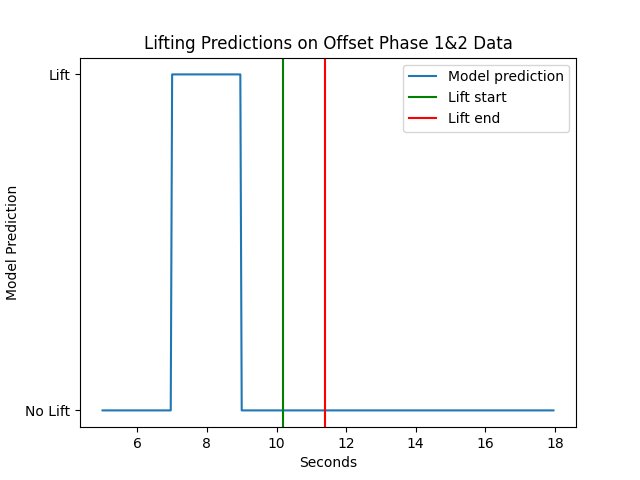}
    \caption{Offset Predictions on Phase 1\&2 data. Model is consistently and systematically predicting several seconds to early, implying a fault in the dataset.}
    \label{fig:offset_predictions}
\end{figure}

Results from the base lab-trained model were promising and produced a very high accuracy \cite{brennans_paper}. When it was discovered the model performed poorly on real-world data, the first step was to perform a detailed examination of the training data (Phase 1\&2). It was discovered that several subjects had systematic offsets between the IMU time and the labeling time. This caused the predictions to be delayed by up to several seconds (see figure \ref{fig:accurate_predictions} and \ref{fig:offset_predictions}). Upon investigation, it appears that the IMU clock drifts out of sync with the camera clock over time. The degree of offset varies from subject to subject, but typically is consistent across trials that occurred back-to-back.\\

This error was less pronounced on the original lab-trained model, likely because the size of the sliding window was 25 frames (1 second) \cite{brennans_paper}. However, when shrinking the size of the sliding window, the issue becomes much more pronounced. The incorrect offsets were adjusted manually and trials without the necessary information to manually fix were removed from the dataset. The model was retrained on Phase 1\&2 and re-evaluated on Phase 3 data. The model showed a noticeable improvement in evaluation and the optimal size of the sliding window changed from 25 frames to 10 frames.\\

\subsubsection{Incorrect Sensor Placement}

Further into the dataset investigation process it was discovered that the back sensor had been incorrectly placed on the subjects' back in several phase 1\&2 trials. Fortunately, the incorrect orientation was consistent, meaning that we could use data from other sensors to infer if the sensor was incorrectly placed and restructure the data mathematically to emulate the correct sensor placement. The resulting sensor data was compared against a correctly placed sensor on the same subject performing a similar motion and was confirmed to be effective.\\

Typically, it would be advisable to remove any abnormal data. However, given our very small dataset we decided to invest time in fixing the data instead of removing it.\\

\subsection{Non-Representative Training Data}
Another challenge we faced porting this model onto real-world data was going from specific laboratory-controlled movement to general real-world movement. The Phase 1\&2 non-lifting dataset consisted of primarily of sitting, standing, walking, and jumping. In the real world, workers are performing much more complex non-lifting motions. This causes the model to struggle to differentiate between these complex movements and a lifting event. For example, the model frequently predicts a lifting event when a worker is utilizing a control panel.\\

Machine learning models learn their classifiers by identifying features that exist when the classifier is present, but don't exist when the classifier is not. If the training data is not fully representative of the deployment environment, the model may assume certain feature combinations always correlate with the classifier when in fact they do not. In our situation, it seems the model highly correlates lifting with a subject having their arms extended out in front of them. This is a reasonable and true correlation in Phase 1\&2 but is not true in the more general Phase 3 data.

\subsection{Poor Model Generalization}
As mentioned before, due to our relatively small and controlled training dataset, our model struggled to interpret the complex motions in phase 3 data. One technique we used to understand \emph{how} the model interpreted the data was saliency mapping. Typically, neural network models are thought of as "black boxes", where input data is fed into the model and outputs are generated. The user has no concept of how the model processes the input data to provide the outputs. This makes it very challenging to debug poorly performing models. Saliency mapping is a technique which uses the weights between neurons as well as an input to create a heatmap of the input features that have the most significance on the output \cite{saliency_mapping_paper}.\\ 

\begin{figure*}
    \centering
    \includegraphics[width=1\textwidth, height=4cm]{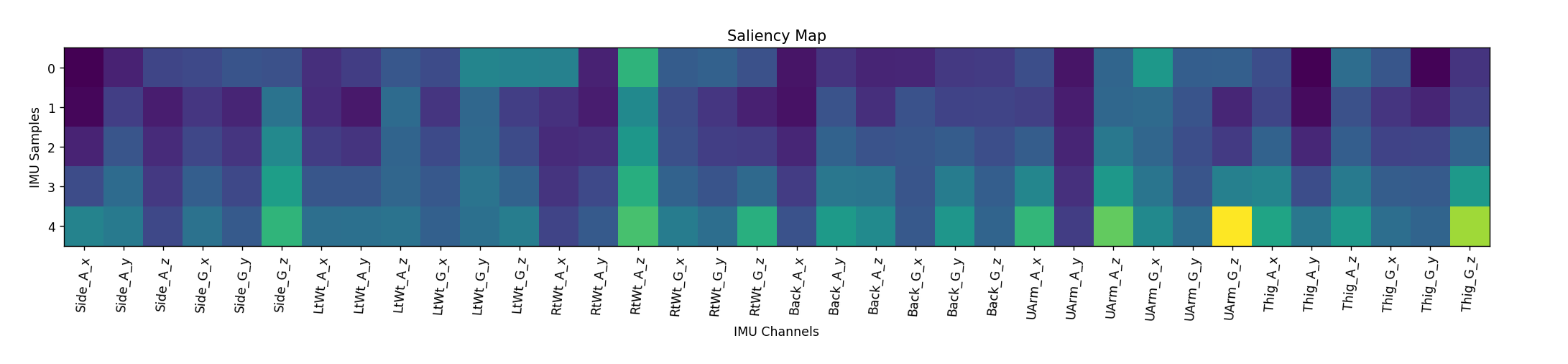}
    \caption{IMU Saliency Map. Lighter colors: higher saliency, darker colors: lower saliency. Columns: IMU channels (gyro or accelerometer) and rows: time steps (1/25th of a second)}
    \label{fig:saliency_IMU_bad}
\end{figure*}

Historically, saliency maps are applied to images where the heatmap can be overlaid on the input image.  However, saliency maps can be used on non-image data as well. For this project, we applied a saliency map to our IMU data. See figure \ref{fig:saliency_IMU_bad}. Each row is a 'frame' in our dataset (representing 1/25th of a second) and each column is a feature from an IMU sensor. The darker squares represent less contribution, while the lighter squares represent more. As we can see in figure \ref{fig:saliency_IMU_bad}, this model uses a single point in our input data as a major indicator of whether a lifting event is occurring or not.\\

Unless we expect a single feature to effectively represent our class, this is typically a bad sign as it implies the model did not generalize well. Here, we can see that the upper arm gyroscope feature is the especially salient feature in our input data. Seeing this should raise a key question "Why is this such a salient feature in our training dataset?" Upon further investigation, we discovered that the only time a subject's hands are out in front of them is when they are lifting. Therefore, using the arm gyro is very effective on the lab data, but as we know, there are frequently times in real-world situations where the arms are extended in front of the body that aren't lifting movements (sitting at a desk, using machinery, etc.).\\

There are several effective ways to resolve this issue. The first and most intuitive is to simply gather additional training data to address this edge case. Unfortunately, gathering additional test data was not financially feasible for this project, so we had to resort to more creative methods. 

\section{Methodology}

With these three challenge-areas identified we investigated four potential solutions to resolve these issues.

\subsection{Synthetic IMU Data}

One idea was to create additional data using lifting data gathered in other public studies. This was originally deemed not feasible as there is no standard placement of IMU sensors. Therefore, the data from other studies utilizing IMUs would not match up with the data we gathered for our study.\\ 

However, many studies utilize Motion Capture data (mocap) to capture the movements of subjects. The AMASS dataset \cite{amass_paper} is a unified database of 15 MoCap datasets containing over 40 hours of motion data. BABEL \cite{babel_paper} builds on the AMASS dataset by labeling frames of each sequence with the action occurring. Additionally, \cite{dip_paper} outlines a methodology to synthesize IMU data from the AMASS dataset by placing virtual IMU sensors on the mesh surface. The combination of frame-by-frame data and synthetic IMU data generation makes this dataset a feasible avenue for incorporating additional data.

\subsection{Transfer Learning}

Another idea was to use transfer learning. Transfer learning is where a model is trained on a large, generalized dataset relating to its final classifier, then is specifically trained on a small amount of data. A common example of this is a voice assistant that responds only to a specific person. The voice assistant is trained on general voice recognition, then later uses a few samples of the user's specific voice to identify that voice uniquely.\\

This idea doesn't work well for IMU data because there are no standard IMU placement locations. Models and datasets that are publicly available do not match our IMU placements, and therefore are not fit for transfer learning in the traditional sense. However, we can still implement a small transfer learning style procedure in this model.\\

Instead of using external data to augment our dataset, we use our lab dataset as the "general" dataset, then a few examples of lifting and non-lifting from a real-world environment. This solution wouldn't fix the lapses in the lab dataset, but it would help the model adjust to any personal bias in lifting technique.

\subsection{Filtering}
Filtering data (or preprocessing) is a common machine learning technique. The idea here was that filtering the data would mask the gaps in our training set and force the model to learn more generally. To do this, we tested two types of filters: Extended Kalman filters and Mahony filters.\\

Extended Kalman filters (EKFs) are a modification of standard Kalman filters, both of which have seen widespread use in IMUs. Both operate fundamentally the same way. Kalman filters first ``predicts'' the current state of the sensor using a state transition model, then ``updates'' using a sensor reading, derived from an observation model. This predict-update cycle allows for handling of sensor noise. In particular, it can assist in the handling of gyroscope drift. EKFs differ from conventional Kalman filters by using non-linear, differentiable functions as part of the model, rather than linear models.\\

An EKF filter is readily available in the AHRS Python library. Mahony filters (available in the same library) were also tested, despite being less widely used, for a more comprehensive evaluation of the performance of filtering.

\subsection{Sensor Removal}
Removing sensors from the dataset seems very un-intuitive. We'd expect that providing the model with more data would allow it to identify correlations that we humans may not see, and while that can be true, in our situation, it actually ends up hurting performance more than helping. With a small dataset, we found that removing features not deemed salient to lifting (by referencing NIOSH equation \cite{niosh_lifting_eq}) improved performance significantly.\\

This process must be done carefully, as removing data limits the capabilities of our model. In our situation, our baseline accuracy was not acceptable for production deployment, so we were willing to sacrifice some sensitivity to gain specificity. It's most important to identify high-risk lifts and since high-risk lifts often have pronounced trunk flexion, we created a set of sensors we felt best captured the motions of high-risk lifting. Models were retrained on their specific sets of sensors and reevaluated on the entire real-world dataset (not just high-risk events).

\section{Results and Discussion}
\label{results}

All results are from the model evaluated on Phase 3 data. The 'Evaluation data' is a subset of the Phase 3 data.

\subsection{Synthetic IMU Data}
The authors of \cite{dip_paper} have provided their tooling for generating synthetic IMU data. However, while the tooling provides a simulation of accelerometer data, gyroscopic data is provided as orientations, requiring modifications to provide angular velocities as in the NIOSH data set. Gyroscopic data is vital in capturing key lifting characteristics; therefore, reconciling the two formats is essential for incorporating additional data. Unfortunately, the necessary modifications proved too far outside our area of experience to incorporate any synthetic data into our training set.

\subsection{Transfer Learning}
Our transfer learning strategy did not produce a significant improvement in accuracy or F1. Typically, transfer learning models are trained on much more data ahead of time and have very good performance on general class detection/identification ahead of the transfer learning stage. In our process, the model did not have a general understanding of what a lifting event was, so finishing training with a small sample of real-world data did not improve performance because the model was already poorly generalized. Simply put, transfer learning works best from taking a powerful generalized model and applying it to a specific sub-set of its original purpose (i.e., written digit recognition). Our model was not a well-generalized model, so our specific transfer training was unsuccessful.

\subsection{Filtering}

\begin{figure}[h]
    \centering
    \includegraphics[width=0.49\textwidth]{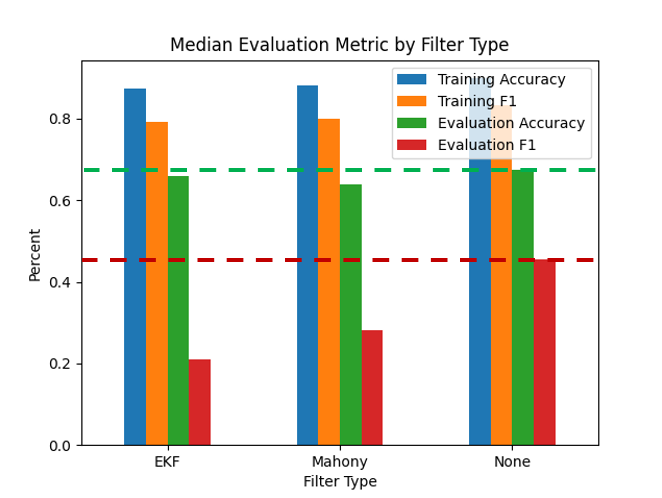}
    \caption{Results from filtering. No filtering provides the highest accuracy and F1 scores on the evaluation dataset}
    \label{fig:filtering_results_bar}
\end{figure}

\begin{figure}[h]
    \centering
    \includegraphics[width=0.49\textwidth]{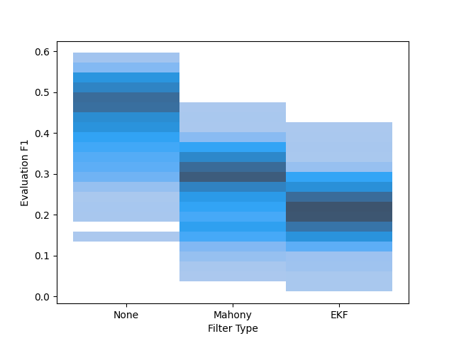}
    \caption{Heatmap results from filtering. There are no models that perform exceptionally better than the mean for any given filter technique}
    \label{fig:filtering_results_heatmap}
\end{figure}

See Figure \ref{fig:filtering_results_bar}. It was found that both EKF and Mahony filters resulted in slightly lower median accuracy and F1 scores on training data, and notably lower median accuracy and F1 scores on evaluation data. Furthermore, Figure \ref{fig:filtering_results_heatmap} shows that even the models with the best F1 scores with filtering applied are outperformed by the median model with no filtering.

It was noted in \cite{brennans_paper} that filtering the NIOSH data set was considered but ultimately not used due to both better performance with raw data and reduced pre-processing time; although it is unclear what filtering methods were evaluated, the results we have obtained seem to align with these preliminary findings.

One additional source of reduced performance from the EKF filter may be from un-optimized filter parameters. Kalman filters incorporate noise variances of the sensors. The default noise levels of the AHRS library were used in testing due to unfamiliarity with the details of the sensors in the NIOSH data set. Similarly, default values were used for gain parameters of the Mahony filter.

\subsection{Sensor Removal}
\begin{figure}[h]
    \centering
    \includegraphics[width=0.49\textwidth]{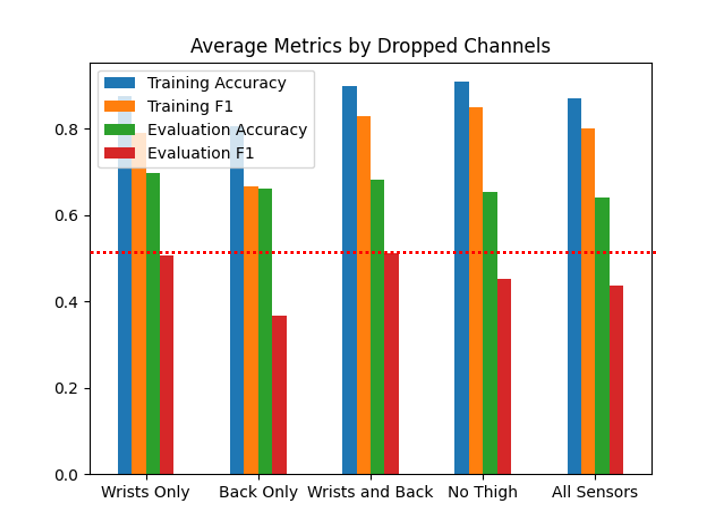}
    \caption{Results from removing sensors. Using only the wrist and back sensors provided the best F1 and accuracy scores on the Phase 3 dataset, showing significant improvement over training with all sensors}
    \label{fig:dropped_sensor_results}
\end{figure}

Removing sensor channels produced excellent results. The best model trained with only wrist and back sensors had an F1 score 8\% higher than the top performing model trained with all sensors. See figure \ref{fig:dropped_sensor_results} for a breakdown of the different core channel combinations and their associated metrics.\\

Hiding data from a Machine Learning model typically does not produce better results, however in this situation it did because the sensors we removed were the \emph{least} representative of real-world lifting motions. In real-world lifting environments, subjects are constantly moving their arms around as part of the normal working environment, however in our training data, non-lifting working motions were not included in the dataset. Therefore, when the model learned our lifting class, it used features in the other sensors that were not unique to lifting generally, but were in our training set. Removing these sensor channels forced the model to learn a more generalized definition of a lifting event from the remaining sensors.

HAR IMU models have come a long way since their original development. We can design models to identify repetitive consistent laboratory-like data reliably using multiple different model architectures. However, the vision of utilizing a laboratory trained lifting model on real-world data is still largely out of reach. Non-standard IMU placements cause limited datasets which make it challenging to incorporate all reasonable movements in training. The authors of  \cite{dip_paper} have successfully emulated IMU sensor data from motion-capture data by placing virtual sensors on models, which could significantly expand the reach of future training data sets. Furthermore, the authors have provided their tooling for generating synthetic IMU data, although modifications may be necessary to reconcile differences between synthetic and real-world data sets. \\

However, even with limited training data, we are still able to achieve significant insight with saliency mapping techniques. We were able to apply saliency mapping to our IMU model and see that the model was fixating on a feature not vitally significant to lifting. From this learning, we were able to target our solutions to resolve this challenge and achieved significant results for our efforts.\\

Our baseline balanced Phase 3 F1 was 32.8\%. By perfecting training data, resolving sensor issues, and performing a targeted removal of non-salient sensors, we were able to increase our typical F1 scores up to ~55\%, with our top model nearly doubling our baseline with a F1 of 64.66\%\\

Unfortunately, even with these substantial increases to performance, we still fall short of the 90+\% necessary for deployment. However, with more work in synthetic data future work could improve performance enough to make this solution viable for deployment.\\

\section{Conclusion}
\label{conclusion}

Outlined below are three key research considerations informed from this project:
\begin{enumerate}
    \item Training Data is very important
    \begin{itemize}
        \item Thoroughly ensure the accuracy of your training set and insure it is representative of your deployment environment. If the dataset isn't representative, it will be very challenging to develop a successful model.
        \item Researching ways to augment datasets: Larger training sets, as long as they're representative, will help your model generalize better. Augmenting data with synthetic data or publicly available datasets is a powerful way to increase your model's robustness.
    \end{itemize}
    \item Studying model's saliency: Having a sense of how your model interprets data is vital to understanding how it will behave in new environments. It can also inform modifications to make ahead of time to improve the chances of success.
    \item Considering transfer learning: Transfer learning is a great way to adjust a model slightly to remove user bias. While transfer learning was unsuccessful on this particular project, transfer learning can provide a "personal touch" to a model that can significantly improve its reliability on a specific user.
\end{enumerate}

\section{Acknowledgments}
This work was funded by CDC/NIOSH. The authors would like to thank Marie Hayden, Menekse Barim and Dwight Werren for their support in creating the datasets. Findings and conclusions in this report are those of the authors and do not necessarily represent the official positions of NIOSH or the Centers for Disease Control and Prevention (CDC). The mention of any company or product does not constitute endorsement by NIOSH or the CDC.

\bibliographystyle{IEEEtran}
\bibliography{bibliography}

\begin{thebibliography}{10}
\providecommand{\url}[1]{#1}
\csname url@samestyle\endcsname
\providecommand{\newblock}{\relax}
\providecommand{\bibinfo}[2]{#2}
\providecommand{\BIBentrySTDinterwordspacing}{\spaceskip=0pt\relax}
\providecommand{\BIBentryALTinterwordstretchfactor}{4}
\providecommand{\BIBentryALTinterwordspacing}{\spaceskip=\fontdimen2\font plus
\BIBentryALTinterwordstretchfactor\fontdimen3\font minus
  \fontdimen4\font\relax}
\providecommand{\BIBforeignlanguage}[2]{{%
\expandafter\ifx\csname l@#1\endcsname\relax
\typeout{** WARNING: IEEEtran.bst: No hyphenation pattern has been}%
\typeout{** loaded for the language `#1'. Using the pattern for}%
\typeout{** the default language instead.}%
\else
\language=\csname l@#1\endcsname
\fi
#2}}
\providecommand{\BIBdecl}{\relax}
\BIBdecl

\bibitem{back_pain_leading_cause_of_injury}
V.~Anderson, B.~Bernard, S.~Burt, L.~Cole, C.~Estill, L.~Fine, K.~Grant,
  C.~Gjessing, L.~Jenkins, J.~Hurrell, N.~Nelson, D.~Pfirman, R.~Roberts,
  D.~Stetson, M.~Haring-Sweeney, and S.~Tanaka, \emph{Musculoskeletal Disorders
  and Workplace Factors: A Critical Review of Epidemiologic Evidence for
  Work-Related Musculoskeletal Disorders of the Neck, Upper Extremity, and Low
  Back}, 01 1997.

\bibitem{poor_workstation_design}
\emph{\BIBforeignlanguage{eng}{Work practices guide for manual lifting.}}, ser.
  DHHS (NIOSH) publication ; no. 81-122.\hskip 1em plus 0.5em minus 0.4em\relax
  Cincinnati, Ohio: U.S. Dept. of Health and Human Services, Public Health
  Service, Centers for Disease Control, National Institute for Occupational
  Safety and Health, Division of Biomedical and Behavioral Science, 1981.

\bibitem{extensively_studied}
\BIBentryALTinterwordspacing
D.~Cook, K.~D. Feuz, and N.~C. Krishnan, ``Transfer learning for activity
  recognition: a survey,'' \emph{Knowledge and Information Systems}, vol.~36,
  no.~3, pp. 537--556, Sep 2013. [Online]. Available:
  \url{https://doi.org/10.1007/s10115-013-0665-3}
\BIBentrySTDinterwordspacing

\bibitem{video_success}
\BIBentryALTinterwordspacing
I.~Rodríguez-Moreno, J.~M. Martínez-Otzeta, B.~Sierra, I.~Rodriguez, and
  E.~Jauregi, ``Video activity recognition: State-of-the-art,'' \emph{Sensors},
  vol.~19, no.~14, 2019. [Online]. Available:
  \url{https://www.mdpi.com/1424-8220/19/14/3160}
\BIBentrySTDinterwordspacing

\bibitem{brennans_paper}
B.~Thomas, M.-L. Lu, R.~Jha, and J.~Bertrand, ``Machine learning for detection
  and risk assessment of lifting action,'' \emph{IEEE Transactions on
  Human-Machine Systems}, vol.~52, no.~6, pp. 1196--1204, 2022.

\bibitem{overfitting_paper}
\BIBentryALTinterwordspacing
X.~Ying, ``An overview of overfitting and its solutions,'' \emph{Journal of
  Physics: Conference Series}, vol. 1168, no.~2, p. 022022, feb 2019. [Online].
  Available: \url{https://dx.doi.org/10.1088/1742-6596/1168/2/022022}
\BIBentrySTDinterwordspacing

\bibitem{niosh_lifting_eq}
\BIBentryALTinterwordspacing
L.~Donisi, G.~Cesarelli, A.~Coccia, M.~Panigazzi, E.~M. Capodaglio, and
  G.~D’Addio, ``Work-related risk assessment according to the revised niosh
  lifting equation: A preliminary study using a wearable inertial sensor and
  machine learning,'' \emph{Sensors}, vol.~21, no.~8, 2021. [Online].
  Available: \url{https://www.mdpi.com/1424-8220/21/8/2593}
\BIBentrySTDinterwordspacing

\bibitem{niosh_labeling_paper}
M.~Barim, M.-L. Lu, S.~Feng, G.~Hughes, M.~Hayden, and D.~Werren, ``Accuracy of
  an algorithm using motion data of five wearable imu sensors for estimating
  lifting duration and lifting risk factors,'' \emph{Proceedings of the Human
  Factors and Ergonomics Society Annual Meeting}, vol.~63, no.~1, p.
  1105–1111, 2019.

\bibitem{saliency_mapping_paper}
K.~Simonyan, A.~Vedaldi, and A.~Zisserman, ``Deep inside convolutional
  networks: Visualising image classification models and saliency maps,'' 2014.

\bibitem{amass_paper}
N.~Mahmood, N.~Ghorbani, N.~F. Troje, G.~Pons-Moll, and M.~J. Black, ``{AMASS}:
  Archive of motion capture as surface shapes,'' in \emph{2019 IEEE/CVF
  International Conference on Computer Vision (ICCV)}, Oct. 2019, pp.
  5441--5450.

\bibitem{babel_paper}
A.~R. Punnakkal, A.~Chandrasekaran, N.~Athanasiou, A.~Quiros-Ramirez, and M.~J.
  Black, ``{BABEL}: Bodies, action and behavior with english labels,'' in
  \emph{Proceedings IEEE/CVF Conf.~on Computer Vision and Pattern Recognition
  (CVPR)}, Jun. 2021, pp. 722--731.

\bibitem{dip_paper}
Y.~Huang, M.~Kaufmann, E.~Aksan, M.~J. Black, O.~Hilliges, and G.~Pons-Moll,
  ``Deep inertial poser learning to reconstruct human pose from sparseinertial
  measurements in real time,'' \emph{ACM Transactions on Graphics, (Proc.
  SIGGRAPH Asia)}, vol.~37, no.~6, pp. 185:1--185:15, Nov. 2018.

\end{thebibliography}

\end{document}